\documentclass{opt2024} 

\title{Intuitive Analysis of the Quantization-based Optimization: From Stochastic and Quantum Mechanical Perspective}

\optauthor{%
\Name{Jinwuk Seok} \Email{jnwseok@etri.re.kr}\\
\Name{Changsik Cho} \Email{cscho@etri.re.kr}\\
\addr{Future Computing Research Division, Artificial Intelligence Research Lab\\
  ETRI\\
  Daejeon, South KOREA, 34129 }
}

\def \bm{\boldsymbol}
\newtheorem{assumption}{Assumption}
\newtheorem{theorem-opt}{Theorem}
\newtheorem{lemma-opt}[theorem-opt]{Lemma}
\SetKwInput{KwInput}{Input}                
\SetKwInput{KwOutput}{Output}
\usepackage{pgf}
\usepackage{verbatim}
\usepackage{booktabs}
\usepackage{listings}
\usepackage{multicols}

\usepackage{pgfplots}
\usepackage{tikz}

\usepackage{multirow}

\begin{document}

\maketitle

\begin{abstract}%
In this paper, we present an intuitive analysis of the optimization technique based on the quantization of an objective function.
Quantization of an objective function is an effective optimization methodology that decreases the measure of a level set containing several saddle points and local minima and finds the optimal point at the limit level set.
To investigate the dynamics of quantization-based optimization, we derive an overdamped Langevin dynamics model from an intuitive analysis to minimize the level set by iterative quantization. We claim that quantization-based optimization involves the quantities of thermodynamical and quantum mechanical optimization as the core methodologies of global optimization.
Furthermore, on the basis of the proposed SDE, we provide thermodynamic and quantum mechanical analysis with Witten-Laplacian. 
The simulation results with the benchmark functions, which compare the performance of the nonlinear optimization, demonstrate the validity of the quantization-based optimization.
\end{abstract}

\section{Introduction}
From a conventional engineering perspective, quantization is one of the significant signal processing techniques, such as effectively compressing original data\cite{Weiss_1979, Ljung_1985, Boutalis_1989}. 
For a long time, the goal of quantization research has been to reduce the quantization error and restore an original signal faithfully from compressed data by quantization. 
Instead of researching quantization as a branch of signal processing, we could find an effective non-convex optimization algorithm that quantizes the level set of an objective function and appropriately decreases the quantization step to time index.
Furthermore, conventional research on quantization demonstrates that if the quantization error does not depend on the original signal when we quantize the signal uniformly for a large amount of data, the quantization error follows an independent increment distribution(i.i.d.)\cite{Zamir_1996, Gray:2006, Marco_2005, Jimnez_2007}.  
Therefore, we can design an effective stochastic optimization algorithm using a suitable quantization process from the i.i.d. quantity of the quantization error. 
For this purpose, we present an intuitive stochastic analysis of an optimization algorithm based on quantization applied to a random search.
The proposed analysis presents the stochastic differential equation (SDE) to describe the dynamics of quantization-based optimization.
We can establish the Witten-Laplacian \cite{Berglund_2011, Bin_JMLR_2023, Liu_2023} to demonstrate that the quantization to an objective function provides the escape property from local minima.  
Finally, we verify the validity of the proposed analysis by comparing the optimization performance of the quantization-based algorithm with that of other conventional global optimization algorithms such as simulated annealing(SA)\cite{SA_83_01, Rere_2015, Lalaoui_2018, Nguyen_2021} and quantum annealing(QA)\cite{Kadowaki_1998, Santoro_2006, Bettina_2015,Biswas_2017}.
\section{Preliminaries}
\subsection{Definition and Assumption}
The conventional research relevant to signal processing defines a quantization such that $x^Q \triangleq \lfloor \frac{x}{\Delta} + \frac{1}{2} \rfloor \Delta$ for $x \in \mathbf{R}$, where $\Delta \in \mathbf{R}^+$ denotes a fixed-value quantization step\cite{Gray:2006, Jimnez_2007}. 
We provide a more detailed definition of quantization to explore the impact of the quantization error, using the quantization parameter as a reciprocal of the quantization step such that $Q_p \triangleq \Delta^{-1}$,  so we denote the quantization parameter as $Q_p$ and the quantization step as $Q_p^{-1}$, respectively.
\begin{definition}  
\label{def_q01}
For ${x} \in \mathbf{R}$, we define the quantization of $x$ as follows:
\begin{equation}
{x}^Q \triangleq \frac{1}{Q_p} \lfloor Q_p \cdot  ({x} + 0.5 \cdot Q_p^{-1}) \rfloor
= \frac{1}{Q_p} \left( Q_p \cdot {x} + {\varepsilon}^q \right) = {x} + {\varepsilon}^q Q_p^{-1}, \quad {x}^Q \in \mathbf{Q},
\label{def_eq01}
\end{equation}
where $\lfloor x \rfloor \in \mathbf{Z}$ denotes the floor function such that $\lfloor x \rfloor \leq x$ for all $x \in \mathbf{R}$, $Q_p \in \mathbf{Q}^+$ denotes the quantization parameter, and $\varepsilon^q$ is the fraction for quantization such that ${\varepsilon}^q : \Omega \mapsto \mathbf{R}[-1/2, 1/2)$. 
\end{definition}
\begin{definition}  
\label{def_02}
The quantization parameter $Q_p$ is a monotone-increasing function such that 
\begin{equation}
Q_p (t) = \eta \cdot b^{\bar{h}(t)}, 
\label{def_eq02}    
\end{equation}
where $\eta \in \mathbf{Q}^{++}$ denotes the fixed constant parameter of the quantization parameter, $b \in \mathbf{Z}^+$ represents the base that is typically $2$, and $\bar{h} :\mathbf{R}^{++} \mapsto \mathbf{Z}^+$ denotes the power function such that $\bar{h}(t) \uparrow \infty \; \text{ as } \; t \rightarrow \infty$.
\end{definition}
\begin{assumption}  
\label{assum_01}
The quantization error $Q_p^{-1}\varepsilon^q$ defined in \eqref{def_eq01} is a uniformly distributed, independent random variable such that  
\begin{equation}
    \mathbb{E}_{\varepsilon^q} Q_p^{-1}\varepsilon^q = 0, \quad \mathbb{E}_{\varepsilon^q} Q_p^{-2} \varepsilon^q = Q_p^{-2} \mathbb{E}_{\varepsilon^q} {\varepsilon^q}^2 =1/(12 \, Q_p^{2})
\end{equation}
\end{assumption}
Furthermore, we can establish an independent stochastic process with a fraction for quantization such as $(\varepsilon_t^q)_{t \geq 0}$. 
Based on Assumption \ref{assum_01}, such a stochastic process is an i.i.d. process, and we can regard the stochastic process $(\varepsilon_t^q)_{t \geq 0}$ as a white noise, which is known as the White Noise Hypothesis (WNH), from conventional researches such as \cite{Zamir_1996, Gray:2006, Marco_2005, Jimnez_2007}.   
In order to proceed with the main discussion, we consider the optimization problem for an objective function $f \in C^{\infty}$ such that
\begin{equation}
    \text{minimize } f:\mathbf{R}^d \mapsto \mathbf{R}^+.
\label{def_eq03}      
\end{equation}
For a combinatorial optimization problem such as the Traveling Salesman Problem(TSP), we deal with an actual input represented as $\bm{x} \in [0, 1]^m$. In such a case, we assume that there exists a proper transformation from a binary input to a real vector space such that $\mathcal{T}:[0, 1]^m \mapsto \mathcal{X} \subset \mathbf{R}^d$, where $\mathcal{X}$ represents the virtual domain of the objective function $f$. 
Consequently, we consider the objective function as $\eqref{def_eq03}$ regardless of whether the domain is related to the problem. 

Finally, we provide the following assumption for the virtual objective function.
\begin{assumption}  
\label{assum_02}
There exists a quadratic virtual objective function $\tilde{f}^Q(\boldsymbol{x}_{t+1})$ such that
\begin{equation}
f^Q(\boldsymbol{x}_{t+1}) - f^Q(\boldsymbol{x}_{t}) 
= \tilde{f}^Q(\boldsymbol{x}_{t+1}) - f^Q(\boldsymbol{x}_{t}). 
\end{equation}
The Hessian of the quadratic virtual function $\tilde{f}(\boldsymbol{x}_t)$ is a positive definite matrix, therein, the maximum eigenvalue $\lambda_0$ of the virtual function satisfies the following:
\begin{equation}
\lambda_0 = \arg_{\lambda} \max_{\boldsymbol{h} \in \mathbf{R}^d} \frac{\lambda}{2} \| \boldsymbol{h}_t \|^2 = \arg_{\lambda} \max_{\boldsymbol{h} \in \mathbf{R}^d} \int_0^1 (1-s) \boldsymbol{h}_t \cdot \nabla_{\boldsymbol{x}}^2 \tilde{f}(\boldsymbol{x}_t + s \boldsymbol{h}_t) \boldsymbol{h}_t ds.
\label{assum_02_eq02}
\end{equation}
\end{assumption}

\subsection{Primitive Analysis of Quantization-based Optimization Algorithm}
\begin{algorithm2e}[bt]
\SetAlgoLined
\DontPrintSemicolon
\caption{Blind Random Search (BRS) with the proposed quantization scheme}\label{alg-1}
\begin{multicols}{2}
    \KwInput{Objective function $f(x) \in \mathbf{R}^{+}$}
    \KwOutput{$x_{opt}, \; f(x_{opt})$}
    \KwData{$x \in \mathbf{R}^n$}
    {\bfseries Initialization} \; 
     $\tau \leftarrow 0$ and $\bar{h}(0) \leftarrow 0$ \;
     Set initial candidate $x_0$ and $x_{opt} \leftarrow x_0$\;
     Compute the initial objective function $f(x_0)$ \;
     Set $b=2$ and $\eta= b^{-\lfloor \log_b (f(x_0) + 1)\rfloor}, \; Q_p \leftarrow \eta$\;
     $f^Q_{opt} \leftarrow \frac{1}{Q_p} \lfloor Q_p \cdot  (f + 0.5 \cdot Q_p^{-1}) \rfloor$\;
    \While{Stopping condition is satisfied}{
         Set $\tau \leftarrow \tau+1$ \;
         Select $x_{\tau}$ randomly and compute $f(x_{\tau})$\;
         $f^Q \leftarrow \frac{1}{Q_p} \lfloor Q_p \cdot  (f + 0.5 \cdot Q_p^{-1}) \rfloor$\;
        \If {$f^Q \leq f^Q_{opt}$ }{     
             $x_{opt} \leftarrow x_{\tau}$ \;
             $\bar{h}(\tau) \leftarrow \bar{h}(\tau) + 1, \; Q_p \leftarrow \eta \cdot b^{\bar{h}(\tau)}$ 
             $f^Q_{opt} \leftarrow \frac{1}{Q_p} \lfloor Q_p \cdot  (f + 0.5 \cdot Q_p^{-1}) \rfloor$ \;
        }    
    }
\end{multicols}
\end{algorithm2e}

We present a fundamental optimization algorithm as Algorithm \ref{alg-1} for combinatorial optimization with a binary domain and a general optimization problem with a continuous domain. 
The presented algorithm is similar to the elementary MCMC algorithm except for the procedure under the if clause, as it is based on a random search employed in SA and QA. 
The most crucial difference is that the presented algorithm quantizes the objective function regarding a randomly selected candidate $x_t$, and the quantization error induced by the quantization adds i.i.d. noise to the original objective function, such that $f^Q(x_t) = f(x_t) + Q_t \varepsilon_t$, where the time index $t$ is equal to $\tau$ in Algorithm $\ref{alg-1}$. 
From the perspective of updating the parameter, this operation is similar to an annealing operation deduced by the acceptance probability in SA and QA.

The other difference is that the algorithm compares the quantized temporary optimal objective function denoted as $f_{\text{opt}}^Q$ with a quantized objective function to the candidate $f^Q(x_t)$. 
Naturally, the quantized function is not a real objective function value, so that we can model a virtual objective function for the quantized objective function.
For instance, when we represent the objective function with a power series based on the given base value $b$ denoted in \eqref{def_eq02} such that $f(x_t) = f_0 \sum_{k=0}^{\infty} c_k b^k$, we can write the quantized objective function as follows:
\begin{equation}
    f(x_t) = f_0 + \sum_{k=1}^{\infty} c_k b^{-k} = f_0 + \sum_{i=1}^{n} c_i b^{-i} + \sum_{j=n+1}^{\infty} c_j b^{-j} = f^Q(x_t) - Q_p^{-1}(t) \varepsilon_t.
\label{prim-eq01}
\end{equation}
In \eqref{prim-eq01}, we can set the quantized objective function such as $f^Q(x_t) = f_0 + \sum_{i=1}^{n} c_i b^{-i}$, and the quantization error as $Q_p^{-1}(t) \varepsilon_t = \sum_{j=n+1}^{\infty} c_j b^{-j}$. 
Therefore, since the quantization step $Q_p^{-1}(t)$ decreases with the power of $b$ as represented in Algorithm \ref{alg-1} and Definition \ref{def_02}, the equation \eqref{prim-eq01} is similar to the Hamiltonian approximation for a tunneling effect in QA\cite{Katzgraber_2015, Biswas_2017, Jinwuk_etrij_2023}.

Moreover, comparing the quantized value of the objective function in the algorithm does not require accurate modeling of the quantization error, so we can design a virtual objective function for the part of the quantization error.  
Notably, instead of using the objective function's accurate Hessian, we can design the virtual function using a simple Hessian for convenient analysis.
This method helps to analyze and establish a learning equation for machine learning with the proposed quantization. 
In the following chapters, we will discuss the design of the virtual objective function in more detail.
\section{Intuitive Analysis of the Objective Quantization}
\subsection{Association to the Stochastic Differential Equation(SDE)}
As shown in Algorithm \ref{alg-1}, the update condition for the temporary optimal point is that $f^Q(\boldsymbol{x}_{t+1}) $ is less than or equal to $f^Q(\boldsymbol{x}_{t})$, that is, $f^Q(\boldsymbol{x}_{t+1}) \leq f^Q(\boldsymbol{x}_{t})$.    
Since $f^Q(\boldsymbol{x}_{t+1}) < f^Q(\boldsymbol{x}_{t})$ is an evident update condition, we do not consider this case. 
Instead, we investigate the condition when the quantized objective function is equal, i.e. $f^Q(\boldsymbol{x}_{t+1}) = f^Q(\boldsymbol{x}_{t})$.

First, we establish a discrete search equation to describe the dynamics of the proposed algorithm regarding the condition of equal quantized objective function. 
Since the algorithm minimizes the objective function, the search equation requires a negative gradient as a primary directional derivative. 
However, as is well known in convex optimization, a positive definite Hessian to the objective function gives information about local minima from the Taylor expansion with the search equation.  
On the other hand, several local minima and saddle points can exist in the sufficiently large domain of the equal quantized objective function. 
Accordingly, the search equation with a negative gradient needs additional components to elaborate the algorithm's dynamics due to the limitation of the Taylor expansion based on a quadratic approximation. 
To compensate for such a limitation, we add a random vector $\bm{r}_t \in \mathbf{R}^d$, which we assume that the expectation is zero, to the candidate of the search equation as follows: 
\begin{equation}
\label{sde-eq01}
\bm{h}_t \triangleq \bm{x}_{t+1} - \bm{x}_t = - \eta \nabla_{\bm{x}} f(\bm{x}_t) + \eta  \bm{r}_t, \quad \eta \in \mathbf{R}(0, 1).  
\end{equation}
\begin{lemma-opt}
\label{lemma_01}      
Given the candidate of the search equation as \eqref{sde-eq01}, suppose that the virtual objective function satisfying Assumption \ref{assum_02} for the condition $f^Q(\boldsymbol{x}_{t+1}) = f^Q(\boldsymbol{x}_{t})$, before the $Q_p(t)$ is updated. Then, the norm of the random vector $\boldsymbol{r}_{t}$ satisfies $\| \boldsymbol{r}_t \| \leq \sqrt{2 \lambda_0 Q_p^{-1}(t)}$ for the update of $\boldsymbol{x}_t$ under the assumption of the quantized objective functions and $\eta = \lambda_0^{-1}$. 
\end{lemma-opt}
We suppose that the algorithm success to update when $f^Q(\boldsymbol{x}_{t+1}) \leq f^Q(\boldsymbol{x}_{t})$ and fail to update when $f^Q(\boldsymbol{x}_{t+1}) >  f^Q(\boldsymbol{x}_{t})$.
Although we cannot estimate the probability of success correctly, we can suppose that the distribution of the norm of $\bm{r}_t$ is symmetric to the condition of $f^Q(\boldsymbol{x}_{t+1}) = f^Q(\boldsymbol{x}_{t})$. 
Therefore, we consider that the variance of $\bm{r}_t$ is equal to $2 \lambda_0 Q_p^{-1}(t) \boldsymbol{I}_d$.
Furthermore, $\boldsymbol{r}_t$ affects primarily the real value of the objective function $f(\boldsymbol{x}_t)$ and not significantly on $f^Q(\boldsymbol{x}_t)$ affected by the quantization step $Q_p^{-1}(t)$.
Accordingly, since $\boldsymbol{r}_t$ is an independent random vector for the quantization error, we can regard $\boldsymbol{r}_t$ as Gaussian white noise without loss of generality. 
Therefore, we obtain the following theorem regarding SDE to describe the proposed search algorithm. 
\begin{theorem-opt}
\label{theorem_02}      
Based on the given candidate \eqref{sde-eq01} of the search equation and the variance of $\bm{r}_t$, i.e., $\mathbb{E} \bm{r}_t \otimes \bm{r}_t = 2 \lambda_0 Q_p^{-1}(t) \bm{I}_d$, we can obtain the approximated SDE form for the proposed quantization-based search algorithm as follows:
\begin{equation}
    d \bm{X}_t = -\nabla_{\bm{x}} f(\bm{X}_t) + \sqrt{2 Q_p^{-1}(t)} d \bm{W}_t,
\label{sde-eq02}    
\end{equation}
where $\bm{W}_t \in \mathbf{R}^d$ denotes a vector valued standard Brownian motion, and $\bm{X}_t\in \mathbf{R}^d$ denotes a random vector corresponding to $\bm{x}_t$ in \eqref{sde-eq01}.
\end{theorem-opt}
The SDE represented as \eqref{sde-eq02} is a standard Langevin dynamics as an overdamped form and not unadjusted Langevin dynamics\cite{Tony_2024_SIAM, Bin_JMLR_2023}. 
Consequently, the proposed search algorithm embeds a Metropolis-Hastings acceptance step controlled by the quantization parameter $Q_p(t)$ so that the provided SDE describes the dynamics of the proposed search algorithm.
\subsection{Quantum Mechanical Quantities of the Quantization-Based Optimization}
In the previous section, we present the search equation formed as the stochastic difference equation \eqref{sde-eq01} and the approximated SDE \eqref{sde-eq02} for the quantization-based optimization. 
Comparing both equations, when we design the noise model that directly affects the directional derivative, the SDE for the search equation denotes the standard SDE employed with a damped variance term represented as a simulated temperature or a quantization step, as we propose. 

An additional advantage of the SDE \eqref{sde-eq02} is that we can analyze the dynamics of the algorithm with quantum mechanics. 
For this purpose, we introduce the Fokker-Plank equation (FPE) for the SDE \eqref{sde-eq02} with a differential operator $\mathcal{L}^*$ as follows \cite{Gardiner_1985, Bin_JMLR_2023, Tony_2024_SIAM}: 
\begin{equation}
\frac{\partial \rho}{\partial t} = \mathcal{L}^* \rho, \quad \mathcal{L}^* \rho = \nabla_{\bm{x}} \cdot (\rho \nabla_{\bm{x}} f(\bm{x}_t)) + Q_p^{-1}(t) \Delta \rho,
\label{quantum-eq01}
\end{equation}
where $\Delta$ denotes the standard Laplacian, $\rho \triangleq \rho(\bm{x}, t) : \mathbf{R}^d \times \mathbf{R}^+ \mapsto \mathbf{R}[0, 1]$ is the density of the random vector $\bm{X}_t$ defined in \eqref{sde-eq02}. The stationary solution of \eqref{quantum-eq01} is the well-known Boltzmann-Gibbs distribution $Z^{-1}\exp(-Q_p(t) f(\bm{x}_t))$, where $Z = \int_{\mathbf{R}^d} \exp (-Q_p(t) f(\bm{x})) d\bm{x} < \infty$\cite{Gardiner_1985}.

For the analysis based on quantum dynamics under the regime of a small quantization step ($Q_p^{-1}(t) \rightarrow 0$), we derive a following Witten-Laplacian on 0-forms $\Delta_{f, h}^{(0)}$ associated with $f$ and the small parameter $h(t) \triangleq 2 Q_p^{-1}(t)$ from the FPE \eqref{quantum-eq01}:
\begin{equation}
    \Delta_{f, h}^{(0)} u = -h^2 \Delta u + (|\nabla_{\bm{x}} f |^2 - h \Delta_{\bm{x}} f) u, \quad e^{f/h} \mathcal{L}^* u e^{-f/h} = -\frac{1}{2h} \Delta_{f, h}^{(0)}u
\label{quantum-eq02}    
\end{equation}
where $u \triangleq u(\bm{x}, t) : \mathbf{R}^d \times \mathbf{R}^+ \mapsto \mathbf{R}$ is the test function defined as $u = \rho \exp(f/h)$.

Substituting $i \hbar/m$ into $h$, and replacing the test function $u$ with $\psi : \mathbf{R}^d \times \mathbf{R} \mapsto \mathbf{C}$, we can obtain the Schrödinger equation from the Witten-Laplacian \eqref{quantum-eq02} and FPE \eqref{quantum-eq01}.
\begin{equation}
i \hbar \frac{\partial \psi}{\partial t} 
= -\frac{\hbar^2}{2m} \Delta_{\bm{x}} \psi - \frac{m}{2} (\| \nabla_{\bm{x}} f \|^2 - \frac{i \hbar}{m}  \Delta_{\bm{x}} f ) \psi 
= -\frac{\hbar^2}{2m} \Delta_{\bm{x}} \psi + V \psi,   
\label{quantum-eq02-01}
\end{equation}
where $V$ denotes a potential energy defined as $V = -\frac{m}{2}(\| \nabla_{\bm{x}} f \|^2 - h \Delta_{\bm{x}} f)$, $m$ denotes a mass of a particle, and $\hbar$ denotes the reduced Plank constant.

To verify the claims that the search algorithm based on the quantized objective function in Chapter 2 involves a quantum mechanical effect, we investigate the potential energy $V$ when the search parameter $\bm{x}_t$ is trapped in a local minimum. 
Under the assumption, we note that the norm of the gradient $\| \nabla_{\bm{x}} f\|$ is zero at the local minimum. 
In spite of the case, the Laplacian term $2Q_p^{-1}(t) \Delta_{\bm{x}} f$ still remains non-zero in $V$; thus, for small $Q_p^{-1}(t) = \frac{h}{2}$, we get
\begin{equation}
\frac{\partial \rho}{\partial t} = \mathcal{L}^* \rho = -\frac{1}{2h} \Delta_{f, h}^{(0)} u \cdot \exp(-f/h) \approx \frac{h}{2} \Delta_{\bm{x}} u \cdot \exp(-f/h) > 0 
\label{quantum-eq03}
\end{equation}
Equation \eqref{quantum-eq03} reveals that the proposed algorithm can increase the probability density $\rho(\bm{x}, t)$ to move the searching point through a potential barrier even if $\bm{x}_t$ falls in a local minimum. Additionally, the stop condition of the algorithm depends on the quantization step $Q_p^{-1}(t)$, which decreases to zero. 
According to the tunneling effect provided by the Schrödinger equation,  the height of the energy barrier that the algorithm can penetrate is equal to the quantization step $Q_p^{-1}(t)$ since the proposed algorithm quantizes the objective function.  

\section{Experimental Result}
\begin{table*}[]
\caption{{Specification of Benchmark Test functions for Performance Test}}
\centering
{\scriptsize
\begin{tabular}{ccc}
\hline
{Function}& {Equation} & optimal point\\ \hline
Xin-She Yang N4     
& $f(x) = 2.0 + (\sum_{i=1}^d \sin^2(x_i) - \exp(-\sum_{i=1}^d x_i^2) \exp(-\sum_{i=1}^d \sin^2 \sqrt{|x_i|})$ 
& $\min f(x) = -1, \; \text{at } x=0$\\
Salomon  
& $f(x) = 1 - \cos \left(2 \pi \sqrt{\sum_{i=1}^d x_i^2} \right) + 0.1 \sqrt{\sum_{i=1}^d x_i^2}$
&$\min f(x) = 0, \; \text{at } x=0$\\
Drop-Wave 
& $f(x) = - \frac{1 + \cos \left( 12  \sqrt{x^2 + y^2} \right)}{0.5 (x^2 + y^2) + 2}$
&$\min f(x) = 0, \; \text{at } x=0$\\
Shaffel N2
&$0.5 + \frac{\sin^2 (x^2 - y^2) - 0.5}{(1 + 0.001(x^2 + y^2)^2}$ 
&$\min f(x) = 0, \; \text{at } x=0$\\
\hline
\end{tabular}
}
\label{result-1}
\end{table*}
\begin{table*}[]
\caption{{Simulation results of standard benchmark test function for nonlinear optimization}}
\centering
{\scriptsize
\begin{tabular}{cccccc}
\hline
{Function}& {Criterion}  & {Simulated Annealing}  & {Quantum Annealing} & {Quantization-Based Optimization} \\ \hline
\multirow{2}{*}{{Xin-She Yang N4}}    
& {Iteration}         & {6420}    & {17*}      & {3144}     \\
& {Improvement ratio} & {54.57\%} & {35.22\%}  & {54.57\%}  \\
\multirow{2}{*}{{Salomon}}  
& {Iteration}         & {1312} & {7092}    & {1727}     \\
& {Improvement ratio} & {99.99\%} & {99.99\%} & {100.0\%}  \\
\multirow{2}{*}{{Drop-Wave}} 
& {Iteration}         & {907} & {3311} & {254}      \\
& {Improvement ratio} & {100.0\%} & {100.0\%} & {100.0\%}  \\
\multirow{2}{*}{{Shaffer N2}}
& {Iteration}         & {7609} & {9657} & {2073}     \\
& {Improvement ratio} & {100.0\%} & {100.0\%} & {100.0\%}  \\ \hline
\end{tabular}
}
\label{result-2}
\end{table*}

We conducted experiments on well-known continuous benchmark functions such as Xin-She Yang N4\cite{Yang_2020}, Salomon\cite{Salomon_2007}, Drop-Wave\cite{Al-Roomi_2015}, and Shaffer N2\cite{SchafferN2_1989} to compare the optimization performance with thermodynamic-based (SA), quantum mechanical (QA) and quantization-based algorithms. 
Since all benchmark test functions are continuous, simulated annealing, quantum annealing, and the proposed quantization-based optimization are enabled to find the global minima within finite iterations.
Furthermore, the quantization-based optimization finds the global minimum within fewer iterations than the SA and QA algorithms.
We predict that the energy barrier induced virtually by the quantization is relatively lower and more easily penetrated than the natural barrier imposed by the object function.  
Such a quantity of the quantization-based search algorithm can improve the optimization performance. 
As for the experiments on the Xin-She-Yang N4 function,  quantum annealing fails to find the global minimum, whereas SA and the proposed algorithm successfully find it. 
The simulation result demonstrates that the presented analysis based on the FPE to thermodynamical and the proposed quantization-based optimization is valid.
\section{Conclusion} 
We present an intuitive analysis of quantization-based optimization based on stochastic analysis and quantum mechanics.
The proposed SDE for the algorithm is a standard overdamped Langevin dynamics appropriate to the algorithm's dynamics.
Based on the presented SDE, we provide an FPE and Witten-Laplacian, including the quantization parameter, to analyze the algorithm's dynamics. 
However, despite the significant performance difference, the analysis formulas presented in this paper are similar to those of SA.
We suspect that the tunneling effect, which we have not investigated sufficiently, is the primary cause. 
In future work, we will analyze the algorithm's dynamics in more detail using a quantum mechanical perspective and combine the quantization of the objective function with the learning equation in machine learning.

\bibliography{neurips_2024}
\newpage

\appendix

\newtheorem{theorem-ack}{Theorem}
\newtheorem{lemma-ack}[theorem-ack]{Lemma}
\section{Acknowledgment}
This work was supported by the Institute for Information and Communications Technology Promotion(IITP) grant funded by the Korean government(MSIP) (2021-0- 00766, Development of Integrated Development Framework that supports Automatic Neural Network Generation and Deployment optimized for Runtime Environment)
\section{Introduction}
We set notation, proof of lemmas and theorems, and more detailed information about the simulation in the manuscript in the following sections.
\section{Notations}
\begin{itemize}
    \item $\mathbf{R}^n$  ~ The n-dimensional space with real numbers
    \item $\mathbf{R}$    ~ $\mathbf{R}^n \vert_{n=1}$
    \item $\mathbf{R}[\alpha, \beta]$ ~ $\{ x \in \mathbf{R} | \alpha \leq  x \leq \beta, \; \alpha, \beta \in \mathbf{R} \}$
    \item $\mathbf{R}(\alpha, \beta]$ ~ $\{ x \in \mathbf{R} | \alpha <  x \leq \beta, \; \alpha, \beta \in \mathbf{R} \}$
    \item $\mathbf{R}[\alpha, \beta)$ ~ $\{ x \in \mathbf{R} | \alpha \leq  x < \beta, \; \alpha, \beta \in \mathbf{R} \}$
    \item $\mathbf{R}(\alpha, \beta)$ ~ $\{ x \in \mathbf{R} | \alpha <  x < \beta, \; \alpha, \beta \in \mathbf{R} \}$
    \item $\mathbf{Q}^n$  ~ The n-dimensional space with rational numbers
    \item $\mathbf{Q}$    ~ $\mathbf{Q}^n \vert_{n=1}$
    \item $\mathbf{Z}$    ~ The 1-dimensional space with integers. 
    \item $\mathbf{N}$    ~ The 1-dimensional space with natural numbers.  
    \item $\mathbf{R}^+$  ~ $\{ x \vert x \geq 0, \;  x \in \mathbf{R}\}$
    \item $\mathbf{R}^{++}$ ~ $\{ x \vert x > 0, \;  x \in \mathbf{R}\}$
    \item $\mathbf{Q}^+$  ~ $\{ x \vert x \geq 0, \;  x \in \mathbf{Q}\}$
    \item $\mathbf{Q}^{++}$ ~ $\{ x \vert x > 0, \;  x \in \mathbf{Q}\}$
    \item $\mathbf{Z}^+$  ~ $\{ x \vert x \geq 0, \;  x \in \mathbf{Z}\}$
    \item $\mathbf{Z}^{++}$ ~ $\{ x \vert x > 0, \;  x \in \mathbf{Z}\}$,  $\mathbf{Z}^{++}$ is equal to $\mathbf{N}$.
    \item $\lfloor x \rfloor$ ~ $\max \{ y \in \mathbf{Z}| y \leq x, \forall x \in \mathbf{R} \}$
    \item $\lceil x \rceil$ ~ $\min \{ y \in \mathbf{Z}| y \geq x, \forall x \in \mathbf{R} \}$
    \item $\nabla_{\bm{x}} f(\bm{x})$ ~ Gradient of the scalar field $f : \mathbf{R}^d \mapsto \mathbf{R}^+$ such that $\nabla_{\bm{x}} : \mathbf{R} \mapsto \mathbf{R}^{d}$. For Euclidean space, $\nabla_{\bm{x}} f(\bm{x}) = \sum_{i=1}^d \frac{\partial f}{\partial x^i} \bm{e}_i$, where $\{\bm{e}_i\}_{i=1}^d =\{\frac{\partial}{\partial x^i}\}_{i=1}^d$ is a local covariant bases
    \item $\nabla_{\bm{x}} \cdot \bm{V}(\bm{x}) $ ~ Divergence of a vector field $\bm{V} : \mathbf{R}^d \mapsto \mathbf{R}^d$ such that $\nabla_{\bm{x}} \cdot : \mathbf{R}^d \mapsto \mathbf{R}$. For Euclidean space, $\nabla_{\bm{x}} \cdot \bm{V}(\bm{x}) = \sum_{i=1}^d \frac{\partial f}{\partial x^i}$.    
    \item $\nabla_{\bm{x}}^2 f(\bm{x})$ ~ Hessian of the scalar field $f : \mathbf{R}^d \mapsto \mathbf{R}^+$ such that $\nabla_{\bm{x}}^2 : \mathbf{R} \mapsto \mathbf{R}^{d \times d}$ computed by $\nabla_{\bm{x}} \nabla_{\bm{x}} f = \sum_{i, j=1}^d \frac{\partial^2 f}{\partial x^i \partial x^j} \bm{e}_i \otimes \bm{e}_j $ for an Euclidean space.
    \item $\Delta_{\bm{x}} f(\bm{x})$ ~ Laplacian of the scalar field $f : \mathbf{R}^d \mapsto \mathbf{R}^+$ such that $\Delta_{\bm{x}} : \mathbf{R} \mapsto \mathbf{R}$ computed by $\Delta_{\bm{x}} f(\bm{x}_t) = \nabla_{\bm{x}} \cdot \nabla_{\bm{x}} f = \sum_{i=1}^d \frac{\partial^2 f}{\partial {x^i}^2}$ for a Euclidean space.
\end{itemize}

\section{Proofs of Theorems in Section 3}
\subsection{Proof of Lemma 1}
\begin{lemma-ack}
Given the candidate of the search equation as \eqref{sde-eq01}, suppose that the virtual objective function which is a quadratic function with a positive Hessian for the condition $f^Q(\boldsymbol{x}_{t+1}) = f^Q(\boldsymbol{x}_{t})$, before the $Q_p(t)$ is updated.  Let the maximum eigenvalue of the Hessian be $\lambda_0$.  Then, the norm of the random vector $\boldsymbol{r}_{t}$ satisfies $\| \boldsymbol{r}_t \| \leq \sqrt{2 \lambda_0 Q_p^{-1}(t)}$ for the update of $\boldsymbol{x}_t$ under the same quantized objective condition. 
\end{lemma-ack}
\begin{proof}
We write the quantization of the objective function $f^Q(\bm{x}_{t+1})$ before the algorithm update the quantization step (i.e., $Q_p^{-1}(t+1) = Q_p^{-1}(t)$) as follows:
\begin{equation*}
\begin{aligned}
f^Q(\boldsymbol{x}_{t+1}) 
&= f(\boldsymbol{x}_{t+1}) + Q_p^{-1}(t+1) \varepsilon_t \\
&= f(\boldsymbol{x}_{t}) + \nabla_{\boldsymbol{x}} f(\boldsymbol{x}_t) \cdot \boldsymbol{h}_{t} + \int_0^1 (1-s) \boldsymbol{h}_{t} \cdot \nabla_{\boldsymbol{x}}^2 f(\boldsymbol{x}_t + s \boldsymbol{h}_t) \boldsymbol{h}_t ds + Q_p^{-1}(t+1) \varepsilon_{t+1}\\
&= f(\boldsymbol{x}_{t}) + Q_p^{-1}(t) \varepsilon_t + \nabla_{\boldsymbol{x}} f(\boldsymbol{x}_t) \cdot \boldsymbol{h}_{t} + \int_0^1 (1-s) \boldsymbol{h}_{t} \cdot \nabla_{\boldsymbol{x}}^2 f(\boldsymbol{x}_t + s \boldsymbol{h}_t) \boldsymbol{h}_t ds \\
&+ Q_p^{-1}(t+1) \varepsilon_{t+1} - Q_p^{-1}(t) \varepsilon_t \\
&= f^Q(\boldsymbol{x}_t) + \nabla_{\boldsymbol{x}} f(\boldsymbol{x}_t) \cdot \boldsymbol{h}_{t} + \int_0^1 (1-s) \boldsymbol{h}_{t} \cdot \nabla_{\boldsymbol{x}}^2 f(\boldsymbol{x}_t + s \boldsymbol{h}_t) \boldsymbol{h}_t ds + Q_p^{-1}(t+1) \varepsilon_{t+1} - Q_p^{-1}(t) \varepsilon_t\\
&= f^Q(\boldsymbol{x}_t) + \nabla_{\boldsymbol{x}} f(\boldsymbol{x}_t) \cdot \boldsymbol{h}_{t} + \int_0^1 (1-s) \boldsymbol{h}_{t} \cdot \nabla_{\boldsymbol{x}}^2 f(\boldsymbol{x}_t + s \boldsymbol{h}_t) \boldsymbol{h}_t ds + Q_p^{-1}(t) (\varepsilon_{t+1} - \varepsilon_t) \\
\end{aligned}    
\end{equation*}

Thus, we obtain the following fundamental equation for the equality condition:
\begin{equation}
f^Q(\boldsymbol{x}_{t+1}) - f^Q(\boldsymbol{x}_t)
= \nabla_{\boldsymbol{x}} f(\boldsymbol{x}_t) \cdot \boldsymbol{h}_{t} + \int_0^1 (1-s) \boldsymbol{h}_{t} \cdot \nabla_{\boldsymbol{x}}^2 f(\boldsymbol{x}_t + s \boldsymbol{h}_t) \boldsymbol{h}_t ds + Q_p^{-1}(t) (\varepsilon_{t+1} - \varepsilon_t)
\label{pf_lm1_eq01}
\end{equation}
To investigate the condition for no imposed random vector such that $\bm{r}_t = 0$, we let the directional derivative as $\bm{h}_t = -\eta \nabla_{\bm{x}} f(\bm{x}_t)$, for $\eta \in \mathbf{R}(0, 1)$. 
Additionally, according to Assumption \ref{assum_02}, we employ a virtual function $\tilde{f}(\bm{x}_t)$ that satisfies $\tilde{f}(\bm{x}_t) = f(\bm{x}_t)$ and $\tilde{f}^Q(\bm{x}_{t+1}) - f^Q(\bm{x}_t) = {f}^Q(\bm{x}_{t+1}) - f^Q(\bm{x}_t)$. 
Hence, we can obtain the following equation: 
\begin{equation}
\begin{aligned}
f^Q(\bm{x}_{t+1}) - f^Q(\bm{x}_t) 
&= \tilde{f}^Q(\bm{x}_{t+1}) - f^Q(\bm{x}_t) \\ 
&= \tilde{f}(\bm{x}_{t+1}) - f(\bm{x}_t) + Q_p^{-1}(t)(\varepsilon_{t+1} - \varepsilon_t), 
\end{aligned}
\label{pf_lm1_eq02}
\end{equation}
where, without loss of generality, we can let the difference of the quantized virtual function and $f^Q(\bm{x}_t)$ such that
\begin{equation}
\tilde{f}(\bm{x}_{t+1}) - f(\bm{x}_t)
= -\nabla_{\bm{x}} f(\bm{x}_t) \cdot \bm{h}_t + \max_{\boldsymbol{h} \in \mathbf{R}^d} \int_0^1 (1 - s) \bm{h}_t \cdot \nabla_{\bm{x}}^2 f(\bm{x}_t) \bm{h}_t ds 
\end{equation}
In \eqref{pf_lm1_eq01} and \eqref{pf_lm1_eq02}, the maximum and minimum of $Q_p^{-1}(\varepsilon_{t+1} - \varepsilon_t)$ is 
\begin{equation}
- Q_p^{-1} = - Q_p^{-1}\left(-\frac{1}{2} - \frac{1}{2} \right) \leq Q_p^{-1}(\varepsilon_{t+1} - \varepsilon_t) \leq Q_p^{-1}\left(\frac{1}{2} + \frac{1}{2} \right) = Q_p^{-1};  
\label{pf_lm1_eq03}
\end{equation}
thus \eqref{pf_lm1_eq03} implies that
\begin{equation}
Q_p^{-1} |\varepsilon_{t+1} - \varepsilon_t| \leq Q_p^{-1}.
\label{pf_lm1_eq04}
\end{equation}

From the assumption of the virtual function \eqref{pf_lm1_eq02}, we can rewrite \eqref{pf_lm1_eq01} as 
\begin{equation*}
\begin{aligned}
f^Q(\bm{x}_{t+1}) - f^Q(\bm{x}_t) 
&= \tilde{f}^Q(\bm{x}_{t+1}) - f^Q(\bm{x}_t) \\ 
&= \tilde{f}(\bm{x}_{t+1}) - f(\bm{x}_t) + Q_p^{-1}(t)(\varepsilon_{t+1} - \varepsilon_t) \\
&= -\nabla_{\bm{x}} f(\bm{x}_t) \bm{h}_t + \frac{1}{2} \lambda_0 \| \bm{h}_t \|^2 + Q_p^{-1}(t)(\varepsilon_{t+1} - \varepsilon_t)\\
&= -\eta \| \nabla_{\bm{x}} f(\bm{x}_t) \|^2 + \frac{1}{2} \lambda_0 \eta^2 \| \nabla_{\bm{x}} f(\bm{x}_t) \|^2 + Q_p^{-1}(t)(\varepsilon_{t+1} - \varepsilon_t)\\ 
&= \frac{1}{2} \lambda_0 \| \nabla_{\bm{x}} f(\bm{x}_t) \|^2 \left(\eta^2 - \frac{2}{\lambda_0} \eta \right) + Q_p^{-1}(t)(\varepsilon_{t+1} - \varepsilon_t)\\
&= \frac{1}{2} \lambda_0 \| \nabla_{\bm{x}} f(\bm{x}_t) \|^2 \left( \left(\eta - \frac{1}{\lambda_0}\right)^2 - \frac{1}{\lambda_0^2} \right) + Q_p^{-1}(t)(\varepsilon_{t+1} - \varepsilon_t).
\end{aligned}
\end{equation*}

Hence, we obtain the following one of a fundamental equation to investigate the update condition:
\begin{equation}
f^Q(\bm{x}_{t+1}) - f^Q(\bm{x}_t) = \frac{1}{2} \lambda_0 \| \nabla_{\bm{x}} f(\bm{x}_t) \|^2 \left( \left(\eta - \frac{1}{\lambda_0}\right)^2 - \frac{1}{\lambda_0^2} \right) + Q_p^{-1}(t)(\varepsilon_{t+1} - \varepsilon_t)
\label{pf_lm1_eq05}
\end{equation}

Since \eqref{pf_lm1_eq05} is a convex with respect to $\eta$, the search algorithm update $\bm{x}_t$ when the minimum of the LHS of \eqref{pf_lm1_eq05} satisfies the following:
\begin{equation}
\min_{\eta \in \mathbf{R}(0, 1)} f^Q(\bm{x}_{t+1}) - f^Q(\bm{x}_t) \leq 0
\label{pf_lm1_eq06}
\end{equation}

Herein, \eqref{pf_lm1_eq06} should satisfy the worst case of quantization error such that $Q_p^{-1}(t)(\varepsilon_{t+1} - \varepsilon_t) = \max_{\varepsilon_{\cdot} \in \mathbf{R}[-0.5, 0.5]} Q_p^{-1}(t)(\varepsilon_{t+1} - \varepsilon_t) = Q_p^{-1}$. 
Considering the above quantization error condition, \eqref{pf_lm1_eq06} implies that
\begin{equation*}
\begin{aligned}
&\min_{\eta \in \mathbf{R}(0, 1)} \frac{1}{2} \lambda_0 \| \nabla_{\bm{x}} f(\bm{x}_t) \|^2 \left( \left(\eta^2 - \frac{1}{\lambda_0}\right)^2 - \frac{1}{\lambda_0^2} \right) + \max_{\varepsilon_{\cdot} \in \mathbf{R}[-0.5, 0.5]} Q_p^{-1}(t)(\varepsilon_{t+1} - \varepsilon_t) \leq 0  \\
&\implies
-\frac{1}{2 \lambda_0}  \| \nabla_{\bm{x}} f(\bm{x}_t) \|^2 + Q_p^{-1} \leq 0.
\implies 
\| \nabla_{\bm{x}} f(\bm{x}_t) \|^2 \geq 2 \lambda_0 Q_p^{-1}
\end{aligned}
\end{equation*}
Therefore, when $\eta = \frac{1}{\lambda_0}$, we obtain the minimum of $\| \nabla_{\bm{x}} f(\bm{x}_t) \|$ such that 
\begin{equation}
\| \nabla_{\bm{x}} f(\bm{x}_t) \| \geq \sqrt{2 \lambda_0 Q_p^{-1}}.
\label{pf_lm1_eq07}
\end{equation}

Next, we consider the case of imposing the random vector $\bm{r}_t$ into the directional derivative $\bm{h}_t$ such that $\bm{h}_t = \eta (-\nabla_{\bm{x}} f(\bm{x}_t) + \bm{r}_t)$. 
Computing the squared norm of $\bm{h}_t$, we get
\begin{equation}
\| \bm{h}_t \|^2 = \eta^2 \| \nabla_{\bm{x}} f(\bm{x}_t) \|^2 - 2 \eta^2 \nabla_{\bm{x}} f(\bm{x}_t) \bm{r}_t + \eta^2 \| \bm{r}_t \|^2 
\label{pf_lm1_eq08}
\end{equation}

Holding the assumption of the virtual function, to avoid the redundant of the quantization error according to the condition of $\| \nabla_{\bm{x}} f(\bm{x}_t)\|$, we substitute \eqref{pf_lm1_eq08} into $\tilde{f}(\bm{x}_{t+1}) - f(\bm{x}_t)$ instead of \eqref{pf_lm1_eq02}; then we can obtain
\begin{equation}
\tilde{f}(\bm{x}_{t+1}) - f(\bm{x}_t)
= \frac{1}{2} \lambda_0 \| \nabla_{\bm{x}} f(\bm{x}_t) \|^2 \left( \left(\eta - \frac{1}{\lambda_0}\right)^2 - \frac{1}{\lambda_0^2} \right)
+ \eta (1 - \eta \lambda_0 ) \nabla_{\bm{x}} f(\bm{x}_t) \cdot \bm{r}_t + \frac{\eta^2}{2} \lambda_0 \| \bm{r}_t \|^2
\label{pf_lm1_eq09}
\end{equation}

Since we have set $\eta$ is equal to $\frac{1}{\lambda_0}$, we can rewrite the RHS of \eqref{pf_lm1_eq09} with the parameter update condition $\tilde{f}(\bm{x}_{t+1}) - f(\bm{x}_t) \leq 0$ such that
\begin{equation}
\tilde{f}(\bm{x}_{t+1}) - f(\bm{x}_t) = 
-\frac{1}{2 \lambda_0} \| \nabla_{\bm{x}} f(\bm{x}_t) \|^2 + \frac{\eta^2}{2} \lambda_0 \| \bm{r}_t \|^2
= \frac{1}{2 \lambda_0} (\| \bm{r}_t \|^2 - \| \nabla_{\bm{x}} f(\bm{x}_t) \|^2 ) \leq 0
\label{pf_lm1_eq10}
\end{equation}
For the minimum of $\| \nabla_{\bm{x}} f(\bm{x}_t) \|^2 $ derived as \eqref{pf_lm1_eq08}, the right most term of \eqref{pf_lm1_eq10} should be less than zero; thus we get
\begin{equation}
\| \bm{r}_t \|^2 \leq \min_{\eta \in \mathbf{R}(0, 1)} \| \nabla_{\bm{x}} f(\bm{x}_t) \|^2 = \sqrt{2 \lambda_0 Q_p^{-1}(t)}
\end{equation}
\end{proof}

\subsection{Proof of Theorem 1}
\begin{theorem-ack}
\label{th_001}      
Based on the given candidate \eqref{sde-eq01} of the search equation and the variance of $\bm{r}_t$, i.e., $\mathbb{E} \bm{r}_t \otimes \bm{r}_t = 2 \lambda_0 Q_p^{-1}(t) \bm{I}_d$, we can obtain the approximated SDE form for the proposed quantization-based search algorithm as follows:
\begin{equation}
    d \bm{X}_t = -\nabla_{\bm{x}} f(\bm{X}_t) + \sqrt{2 Q_p^{-1}(t)} d \bm{W}_t,
\label{sde-ack-eq02}    
\end{equation}
where $\bm{W}_t \in \mathbf{R}^d$ denotes a vector valued standard Brownian motion, and $\bm{X}_t\in \mathbf{R}^d$ denotes a random vector corresponding to $\bm{x}_t$ in \eqref{sde-eq01}.
\end{theorem-ack}
\begin{proof}
Let the time index $\tau_t$ depend on $\eta$ be such that $\tau_t \triangleq t \eta$, where $t$ denotes the time index in \eqref{sde-eq01}. 
Furthermore, we let a vector-valued random variable $\bm{X}_{\tau_t} \in \mathbf{R}^d$ corresponding to $\bm{x}$ in \eqref{sde-eq01}. 
The first-order Taylor expansion of the standard Wiener process is as follows:
\begin{equation}
\bm{W}_{\tau_{t+1}} - \bm{W}_{\tau_{t}} = \eta \frac{d \bm{W}_{\tau_{t}}}{d \tau} + \mathcal{O}(\eta^2),
\label{th_pf_eq01}
\end{equation}
and the variance of the LHS in \eqref{th_pf_eq01} is
\begin{equation}
\mathbb{E}(\bm{W}_{\tau_{t+1}} - \bm{W}_{\tau_{t}}) \otimes (\bm{W}_{\tau_{t+1}} - \bm{W}_{\tau_{t}}) = (\tau_{t+1} - \tau_t)\bm{I}_d = (t + 1 - t)\eta \bm{I}_d = \eta \bm{I}_d. 
\label{th_pf_eq02}  
\end{equation}
From the assumption for $\bm{r}_{\tau_t}$ as a Gaussian white noise and \eqref{th_pf_eq02}, we get
\begin{equation}
\begin{aligned}
\mathbb{E} \bm{r}_{\tau_t} \otimes \bm{r}_{\tau_t} 
= 2 \lambda_0 Q_p^{-1}(\tau_t) \bm{I}_d 
&= 2 \lambda_0 Q_p^{-1}(\tau_t) \mathbb{E}\frac{(\bm{W}_{\tau_{t+1}} - \bm{W}_{\tau_{t}}) \otimes (\bm{W}_{\tau_{t+1}} - \bm{W}_{\tau_{t}})}{\eta} \\
&= 2 \lambda_0^2 Q_p^{-1}(\tau_t) \mathbb{E}\frac{(\bm{W}_{\tau_{t+1}} - \bm{W}_{\tau_{t}}) \otimes (\bm{W}_{\tau_{t+1}} - \bm{W}_{\tau_{t}})}{\eta}. 
\end{aligned}
\label{th_pf_eq03}  
\end{equation}
Therefore, from \eqref{th_pf_eq03}, we can rewrite the random vector $\bm{r}_t$ such that
\begin{equation}
    \bm{r}_{\tau_t} = \lambda_0 \sqrt{2 Q_p^{-1}(\tau_t)} (\bm{W}_{\tau_{t+1}} - \bm{W}_{\tau_{t}}).
    \label{th_pf_eq04}  
\end{equation}
Rewriting \eqref{sde-eq01} to the time index $\tau_t$ and $\bm{X}_{\tau_t}$, we can obtain 
\begin{equation}
\bm{X}_{\tau_{t+1}} - \bm{X}_{\tau_{t}} = - \eta \nabla_{\bm{x}} f(\bm{X}_{\tau_{t}}) + \eta  \bm{r}_{\tau_{t}}.
\label{th_pf_eq05}  
\end{equation}
Substituting \eqref{th_pf_eq04} into $\bm{r}_{\tau_t}$ in \eqref{th_pf_eq05}, we get 
\begin{equation*}
\begin{aligned}
&\bm{X}_{\tau_{t+1}} - \bm{X}_{\tau_{t}} = - \eta \nabla_{\bm{x}} f(\bm{X}_{\tau_{t}}) + \eta \cdot \lambda_0 \sqrt{2 Q_p^{-1}(\tau_t)}(\bm{W}_{\tau_{t+1}} - \bm{W}_{\tau_{t}})\\
&\implies
\bm{X}_{\tau_{t+1}} - \bm{X}_{\tau_{t}} = - \eta \nabla_{\bm{x}} f(\bm{X}_{\tau_{t}}) + \sqrt{2 Q_p^{-1}(\tau_t)}\left( \eta \frac{d \bm{W}_{\tau_{t}}}{d \tau} + \mathcal{O}(\eta^2) \right)\\
&\implies
\frac{\bm{X}_{\tau_{t+1}} - \bm{X}_{\tau_{t}}}{\eta} 
= - \nabla_{\bm{x}} f(\bm{X}_{\tau_{t}}) + \sqrt{2 Q_p^{-1}(\tau_t)}\left(\frac{d \bm{W}_{\tau_{t}}}{d \tau} + \mathcal{O}(\eta) \right) \\
&\implies
\lim_{\eta \rightarrow 0} \frac{\bm{X}_{\tau_{t+1}} - \bm{X}_{\tau_{t}}}{(t+1 - t)\eta} 
= - \nabla_{\bm{x}} f(\bm{X}_{\tau_{t}}) + \sqrt{2 Q_p^{-1}(\tau_t)}\left(\frac{d \bm{W}_{\tau_{t}}}{d \tau} + \lim_{\eta \rightarrow 0} \mathcal{O}(\eta) \right) \\
&\implies
\frac{d \bm{X}_{\tau_{t}}}{d \tau} = - \nabla_{\bm{x}} f(\bm{X}_{\tau_{t}}) + \sqrt{2 Q_p^{-1}(\tau_t)} \frac{d \bm{W}_{\tau_{t}}}{d \tau}.
\end{aligned}
\end{equation*}
Replace the time index $\tau_{t}$ with $t$, it implies that
\begin{equation}
d \bm{X}_{\tau_{t}} = -\nabla_{\bm{x}} f(\bm{X}_{\tau_{t}}) dt + \sqrt{2 Q_p^{-1}(\tau_t)} d \bm{W}_t.   
\end{equation}
\end{proof}
\subsection{From Fokker-Plank Equation to Schrödinger Equation with Witten Laplacian}
Given the Fokker-Plank Equation (FPE) as 
\begin{equation}
\frac{\partial \rho}{\partial t} = \mathcal{L}^* \rho, \quad \mathcal{L}^* \rho = \nabla_{\bm{x}} \cdot (\rho \nabla_{\bm{x}} f(\bm{x}_t)) + Q_p^{-1}(t) \Delta \rho,
\end{equation}
there are two approaches for the derivation of  Witten-Laplacian; one is based on the Fokker-Plank differential operator $\mathcal{L}^*$ as presented in the paper, and the other is using a non-negative symmetric differential $d_{f, h}^{(0)}$ and $d_{f, h}^{(0)*}$, where  the differential denotes as follows \cite{LePeutrec_2021, Liu_2023, Tony_2024_SIAM}:
\begin{equation}
\begin{aligned}
    d_{f, h}^{(0)} &= \exp(-f/h) (h \nabla_{\bm{x}}) \exp(f/h) \\
    d_{f, h}^{(0)*}&= -\exp(f/h) (h \nabla_{\bm{x}} \cdot) \exp(-f/h).
\label{qack-eq01}    
\end{aligned}
\end{equation}
Under the differential defined as \eqref{qack-eq01}, the Witten-Laplacian is as follows:
\begin{equation}
    \Delta_{f, h}^{(0)} = d_{f, h}^{(0)*} d_{f, h}^{(0)}
    \label{qack-eq01-01}   
\end{equation}
We present both derivations in this section. 
First, we derive the Whitten-Laplacian based on the FPE operator as presented \eqref{quantum-eq01}.
We rewrite the FPE as follows
\begin{equation}
\frac{\partial \rho}{\partial t} 
= \nabla_{\bm{x}}(\rho \cdot \nabla_{\bm{x}} f) + Q_p^{-1} \Delta \rho 
= \nabla_{\bm{x}} \rho \cdot \nabla_{\bm{x}} f + \rho \Delta_{\bm{x}} f + \frac{h}{2} \Delta_{\bm{x}} \rho,
\label{qack-eq02}    
\end{equation}
where $h \triangleq 2 Q_p^{-1}$, and $\rho \triangleq \rho(\bm{x}, t) : \mathbf{R}^d \times \mathbf{R}^+ \mapsto \mathbf{R}[0, 1]$ is the density of the random vector $\bm{X}_t$ defined in \eqref{sde-eq02}. The stationary solution of $\rho$ is the Boltzmann-Gibbs distribution $Z^{-1}\exp(-Q_p(t) f(\bm{x}_t))$, where $Z = \int_{\mathbf{R}^d} \exp (-Q_p(t) f(\bm{x})) d\bm{x} < \infty$\cite{Gardiner_1985}.

Let a wave function be $\psi : \mathbb{R}^d \times \mathbb{R} \rightarrow \mathbb{C}$ such that $\rho(\boldsymbol{x}, t) \triangleq \psi(\boldsymbol{x}, t) \exp(-f/h) = \psi g$.
Additionally, we let the fundamental partial derivatives of $\rho$ to $t$ be as follows:
\begin{equation}
    \frac{\partial \rho}{\partial t} = -g \frac{\partial \psi}{\partial t}.
    \label{qack-eq03}
\end{equation}
The gradient of $\rho$ is 
\begin{equation}
    \nabla_{\bm{x}} \rho = \nabla_{\bm{x}} (\psi \cdot g) = \nabla_{\bm{x}} \psi \cdot g - \psi \cdot \frac{\nabla_{\bm{x}} f}{h} g = \left( \nabla_{\bm{x}} \psi - \frac{\psi}{h} \nabla_{\bm{x}} f \right)g.
    \label{qack-eq04}
\end{equation}
The Laplacian of $\rho$ is
\begin{equation}
\begin{aligned}
\Delta_{\bm{x}} \rho 
&= \nabla_{\bm{x}} \cdot \left( \nabla_{\bm{x}} \psi - \frac{\psi}{h} \nabla_{\bm{x}} f \right)g \\
&= \left( \Delta_{\bm{x}} \psi - \frac{1}{h}(\nabla_{\bm{x}} \psi \cdot \nabla_{\bm{x}} f + \psi \Delta_{\bm{x}} f) + \left( \nabla_{\bm{x}} \psi - \frac{\psi}{h} \nabla_{\bm{x}} f \right) \frac{-\nabla_{\bm{x}} f}{h}  \right) g \\
&= \left( \Delta_{\bm{x}} \psi - \frac{2}{h} \nabla_{\bm{x}} \psi \cdot \nabla_{\bm{x}} f - \frac{1}{h} \psi \Delta_{\bm{x}} f + \frac{\psi}{h^2} \| \nabla_{\bm{x}} f\|^2 \right)g.
\label{qack-eq05}
\end{aligned}
\end{equation}
\eqref{qack-eq05} implies that
\begin{equation}
\frac{h}{2} \Delta_{\bm{x}} \rho 
= \left( \frac{h}{2} \Delta_{\bm{x}} \psi -\nabla_{\bm{x}} \psi \cdot \nabla_{\bm{x}} f + \frac{1}{2h} (\| \nabla_{\bm{x}} f \|^2 - h \Delta_{\bm{x}} f) \psi \right)g.
\end{equation}
Furthermore,
\begin{equation}
\begin{aligned}
\nabla_{\bm{x}} (\rho \cdot \nabla_{\bm{x}} f) 
&= \nabla_{\bm{x}} \rho \cdot \nabla_{\bm{x}} f + \rho \Delta_{\bm{x}} f \\
&= \left( \left( \nabla_{\bm{x}} \psi - \frac{\psi}{h} \nabla_{\bm{x}} f \right) \cdot \nabla_{\bm{x}} f + \psi \Delta_{\bm{x}} f \right) g\\
&= \left( \nabla_{\bm{x}} \psi \cdot \nabla_{\bm{x}} f - \frac{1}{h}(\| \nabla_{\bm{x}} f \|^2 - h \Delta_{\bm{x}} f) \psi \right)g
\end{aligned}
\label{qack-eq06}
\end{equation}
Substituting \eqref{qack-eq06} and \eqref{qack-eq05} into \eqref{qack-eq02}, we obtain
\begin{equation}
\begin{aligned}
-g \frac{\partial \psi}{\partial t} 
&= \left( \nabla_{\bm{x}} \psi \cdot \nabla_{\bm{x}} f - \frac{1}{h}(\| \nabla_{\bm{x}} f \|^2 - h \Delta_{\bm{x}} f) \psi + \frac{h}{2} \Delta_{\bm{x}} \psi -\nabla_{\bm{x}} \psi \cdot \nabla_{\bm{x}} f + \frac{1}{2h} (\| \nabla_{\bm{x}} f \|^2 - h \Delta_{\bm{x}} f) \psi \right)g \\
\frac{\partial \psi}{\partial t} 
&= -\frac{h}{2} \Delta_{\bm{x}} \psi + \frac{1}{2h} (\| \nabla_{\bm{x}} f \|^2 - h \Delta_{\bm{x}} f) \psi 
= \Delta_{f, h}^{(0)} \psi.
\end{aligned}
\label{qack-eq07}
\end{equation}
To obtain the Schrödinger equation, we replace $h$ with $i\hbar/m$ in the final equation of \eqref{qack-eq07}.
Substituting the conjugate wave function $\psi^*$ such that $\frac{\partial \rho}{\partial t} = g \frac{\partial \psi^*}{\partial t}$ into \eqref{qack-eq07}, we obtain the following:
\begin{equation*}
\begin{aligned}
g \frac{\partial \psi^*}{\partial t} 
&= \left( \nabla_{\bm{x}} \psi^* \cdot \nabla_{\bm{x}} f - \frac{1}{h}(\| \nabla_{\bm{x}} f \|^2 - h \Delta_{\bm{x}} f) \psi^* + \frac{h}{2} \Delta_{\bm{x}} \psi^* -\nabla_{\bm{x}} \psi^* \cdot \nabla_{\bm{x}} f + \frac{1}{2h} (\| \nabla_{\bm{x}} f \|^2 - h \Delta_{\bm{x}} f) \psi^* \right)g \\
\frac{\partial \psi^*}{\partial t} 
&= \frac{h}{2} \Delta_{\bm{x}} \psi^* - \frac{1}{2h} (\| \nabla_{\bm{x}} f \|^2 - h \Delta_{\bm{x}} f) \psi^* \\
&\implies 
\frac{\partial \psi^*}{\partial t} = \frac{i \hbar}{2m} \Delta_{\bm{x}} \psi^* - \frac{m}{2 i \hbar}(\| \nabla_{\bm{x}} f \|^2 - h \Delta_{\bm{x}} f) \psi^* \\
&\implies 
i \hbar \frac{\partial \psi^*}{\partial t} = -\frac{\hbar^2}{2m} \Delta_{\bm{x}} \psi^* - \frac{m}{2}(\| \nabla_{\bm{x}} f \|^2 - h \Delta_{\bm{x}} f) \psi^* \\
\end{aligned}
\end{equation*}

Consequently, we can obtain the Schrödinger equation \eqref{quantum-eq02-01} replacing $\psi^*$ with $\psi$ as follows:
\begin{equation}
i \hbar \frac{\partial \psi}{\partial t} = -\frac{\hbar^2}{2m} \Delta_{\bm{x}} \psi + V \psi,   
\label{qack-eq08}
\end{equation}
where $V = -\frac{m}{2}(\| \nabla_{\bm{x}} f \|^2 - h \Delta_{\bm{x}} f)$.

The second approach is the methodology using the symmetric differential defined as \eqref{qack-eq01}.
We calculate the first differential $d_{f, h}^{(0)}$ such that
\begin{equation*}
\begin{aligned}
d_{f, h}^{(0)} u
&= e^{-f/h} (h \nabla_{\bm{x}} ) e^{f/h} \, u \\
&= e^{-f/h} \left(h \nabla_{\bm{x}} e^{f/h} \, u + h \, e^{f/h} \, \nabla_{\bm{x}} u \right) \\
&= e^{-f/h} \left(u\, h\, \frac{1}{h} \nabla_{\bm{x}} f\, e^{f/h} + h e^{f/h} \nabla_{\bm{x}} u \right) \\
&= \left( \nabla_{\bm{x}} f + h \nabla_{\bm{x}} \right)u. 
\end{aligned}
\end{equation*}
Next, we calculate the second differential $d_{f, h}^{(0)^*}$ such that
\begin{equation*}
\begin{aligned}
d_{f, h}^{(0)*} \nabla_{\bm{x}} u
&= -e^{f/h} (h \nabla_{\bm{x}} \cdot) e^{-f/h} \nabla_{\bm{x}} u \\
&= -e^{f/h} h (\nabla_{\bm{x}} e^{-f/h} \cdot \nabla_{\bm{x}} u + e^{-f/h} \nabla_{\bm{x}} \cdot \nabla_{\bm{x}} u) \\
&= -e^{f/h} h \left( -\frac{1}{h} e^{-f/h} \nabla_{\bm{x}} f \cdot \nabla_{\bm{x}} u + e^{-f/h} \Delta_{\bm{x}} u\right)\\
&= (\nabla_{\bm{x}} f \cdot \nabla_{\bm{x}} - h \Delta_{\bm{x}}) u
\end{aligned}
\end{equation*}
Consequently, we get
\begin{equation}
\begin{aligned}
d_{f, h}^{(0)} u         &= (\nabla_{\bm{x}} f + h \nabla_{\bm{x}}) u &\because d_{f, h}^{(0)} : \mathbf{R} \rightarrow \mathbf{R}^d\\
d_{f, h}^{(0)*} \nabla_{\bm{x}} u &= (\nabla_{\bm{x}} f - h \nabla_{\bm{x}}) \cdot \nabla_{\bm{x}} u &\because d_{f, h}^{(0)*} : \mathbf{R}^d \rightarrow \mathbf{R}
\end{aligned}
\label{qack-eq09}
\end{equation}
Therefore,
\begin{equation*}
\begin{aligned}
d_{f, h}^{(0)*} d_{f, h}^{(0)} u 
&= (\nabla_{\bm{x}} f - h \nabla_{\bm{x}}) \cdot (\nabla_{\bm{x}} f + h \nabla_{\bm{x}}) u \\
&= (\nabla_{\bm{x}} f - h \nabla_{\bm{x}}) \cdot (u \nabla_{\bm{x}} f + h \nabla_{\bm{x}} u)  \\
&= u \nabla_{\bm{x}} f \cdot \nabla_{\bm{x}} f + h \nabla_{\bm{x}} f \cdot \nabla_{\bm{x}} u - h \nabla_{\bm{x}} \cdot (u \nabla_{\bm{x}} f) -h^2 \nabla_{\bm{x}} \cdot \nabla_{\bm{x}} u \\
&= \| \nabla_{\bm{x}} f \|^2 u + h \nabla_{\bm{x}} f \cdot \nabla_{\bm{x}} u - h \nabla_{\bm{x}} u \cdot \nabla_{\bm{x}} f - h u \nabla_{\bm{x}} \cdot \nabla_{\bm{x}} f -h^2 \Delta_{\bm{x}} u \\
&= \| \nabla_{\bm{x}} f \|^2 u + h (\nabla_{\bm{x}} f \cdot \nabla_{\bm{x}} u - \nabla_{\bm{x}} u \cdot \nabla_{\bm{x}} f) - h \Delta_{\bm{x}} f u - h^2 \Delta_{\bm{x}} u \\
&= -h^2 \Delta_{\bm{x}} u + (\| \nabla_{\bm{x}} f \|^2 - h \Delta_{\bm{x}} f)u
\end{aligned}
\end{equation*}
It implies that 0-form Witten-Laplacian from the computation of the symmetric differentials as follows:
\begin{equation}
    \Delta_{f, h}^{(0)} u = -h^2 \Delta_{\bm{x}} u + (\| \nabla_{\bm{x}} f \|^2 - h \Delta_{\bm{x}} f)u
    \label{qack-eq10}
\end{equation}

We can obtain the Schrödinger equation \eqref{qack-eq08} by applying the Witten-Laplacian presented as \eqref{qack-eq10} to \eqref{qack-eq07} replacing $h$ with $i\hbar/m$ for the conjugate wave function $\psi^*$ instead of $u$.

\subsection{Intuitive Analysis at Local Minima of an Objective Function}
\begin{figure}
    \centering
    \includegraphics[width=0.7\textwidth]{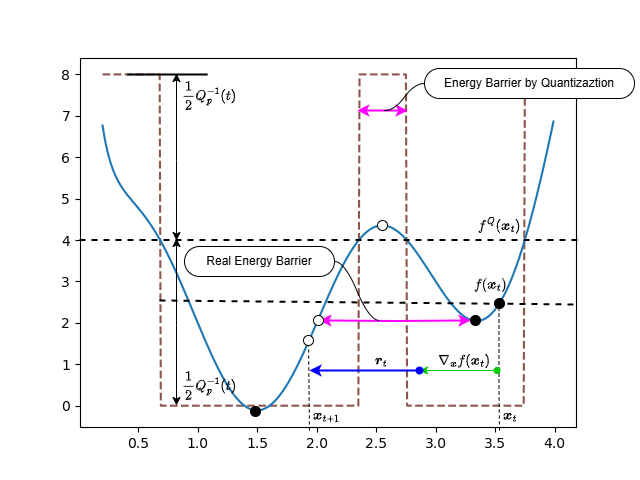}
    \caption{Conceptual diagram of the tunneling effect provided by the quantization; Blue line represents the objective function $f(\bm{x}_t)$, dashed line represents the quantized objective function $f^Q(\bm{x}_t$ with a relatively large quantization step}
    \label{fig:pwp_orif01}
\end{figure}

 We can derive the equation \eqref{quantum-eq03} straightforwardly with \eqref{quantum-eq01} and \eqref{quantum-eq02}.
Therefore, expanding the Witten-Laplacian, we can get the partial derivative of the probability density $\rho$ as follows:
\begin{equation}
\frac{\partial \rho}{\partial t} 
= -\frac{1}{2h} \left( -h^2 \Delta_{\bm{x}} + (\| \nabla_{\bm{x}} f \|^2 - h \Delta_{\bm{x}} f) \right) u \exp(-f/h)
\label{qack-eq11}
\end{equation}

Suppose that the parameter $\bm{x}_t$ remains a local minima $\bm{x}_*$, and the assumption implies that $\nabla_{\bm{x}} f(\bm{x}_*) = 0$ at $t \in \mathbf{R}^+$.  
Particularly, from the fact that $h = 2Q_p^{-1}$, we assume that $h$ is sufficiently small to neglect the effect of $h \Delta_{\bm{x}} u$.

Holding such the assumption for paralysis of searching caused by local minima, we note that there exists a non-zero term $\Delta_{\bm{x}} f \rho$ in the RHS of the Witten-Laplacian \eqref{qack-eq11}. 
Since the trapping point is a local minimum, the eigenvalue of the Laplacian $\Delta_{\bm{x}} f$ is positive, and it implies that the probability density represents increasing as follows:
\begin{equation}
\begin{aligned}
\frac{\partial \rho}{\partial t} 
&= \frac{h}{2} \Delta_{\bm{x}} \rho - \frac{1}{2h} \| \nabla_{\bm{x}} f \|^2 + \frac{1}{2}\Delta_{\bm{x}} f \cdot \rho \\
&= \left( \frac{h}{2} \Delta_{\bm{x}} u -\nabla_{\bm{x}} u \cdot \nabla_{\bm{x}} f + \frac{1}{2h} \| \nabla_{\bm{x}} f \|^2 - \frac{1}{2} \Delta_{\bm{x}} f \cdot u \right) \exp(-f/h) + \frac{1}{2} \Delta_{\bm{x}} f \cdot \rho \\
&= \frac{h}{2} \Delta_{\bm{x}} u \cdot \exp(-f/h) - \frac{1}{2} \Delta_{\bm{x}} f \cdot \rho  + \frac{1}{2} \Delta_{\bm{x}} f \cdot \rho, \quad  \because \nabla_{\bm{x}} f = 0  \\
&= \frac{h}{2} \Delta_{\bm{x}} u \cdot \exp(-f/h) > 0, 
\end{aligned}
\end{equation}
Therefore, the algorithm's stop condition is that the quantization step $h(t) = 2 Q_p^{-1}(t)$ decreases infinitely to zero, which implies that the probability density reaches a stationary distribution.   

\subsection{Specification of the Benchmark Functions in Simulation Results}
\begin{figure}[htb!] 
    \centering
    \subfigure[3D plot of Xin She Yang N4 benchmark function on 2-dimensional input space]{\resizebox{0.48\textwidth}{!}{\includegraphics[width=0.8\textwidth]{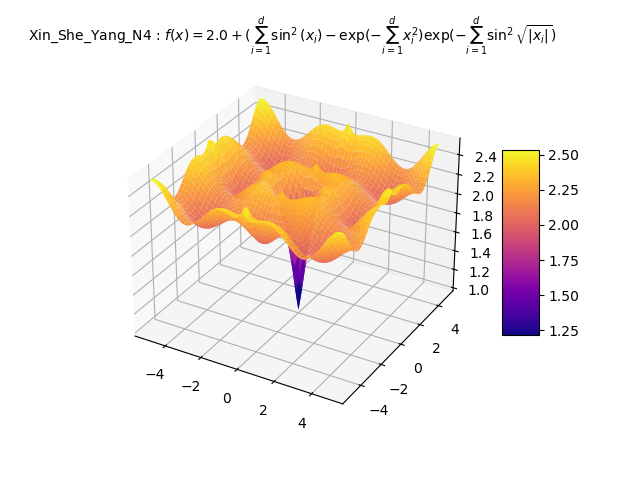}}}\label{fig01-000}
    \hfill
    \subfigure[3D plot of Drop wave benchmark function on 2-dimensional input space]{\resizebox{0.48\textwidth}{!}{\includegraphics[width=0.8\textwidth]{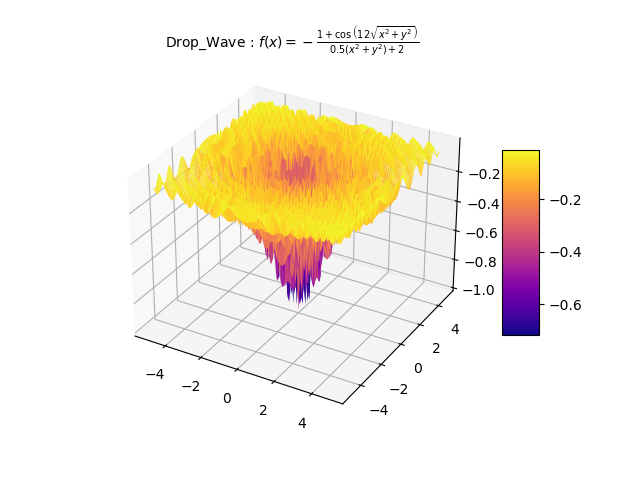}}}\label{fig01-001}
    \hfill
    \subfigure[3D plot of Salomon benchmark function on 2-dimensional input space]{\resizebox{0.48\textwidth}{!}{\includegraphics[width=0.8\textwidth]{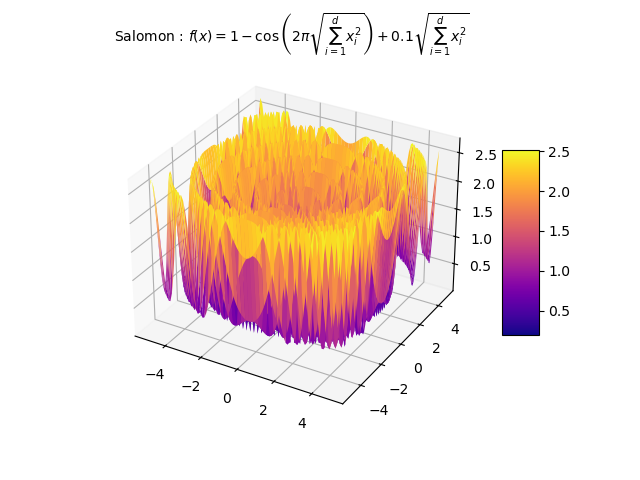}}}\label{fig01-002}
    \hfill
    \subfigure[3D plot of Salomon benchmark function on 2-dimensional input space]{\resizebox{0.48\textwidth}{!}{\includegraphics[width=0.8\textwidth]{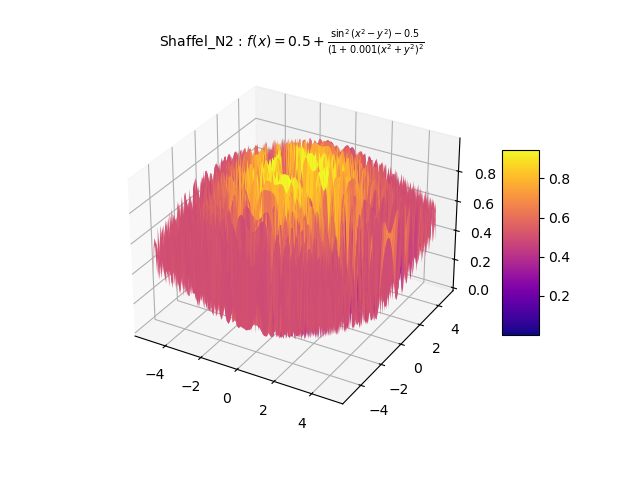}}}\label{fig01-003}
\caption{3D plot of the benchmark functions}
\label{fig_tsp_01}
\end{figure}

\begin{figure}[htb!] 
    \centering
    \subfigure[Search process with respect to Xin She Yang N4 function]{\resizebox{0.48\textwidth}{!}{\input{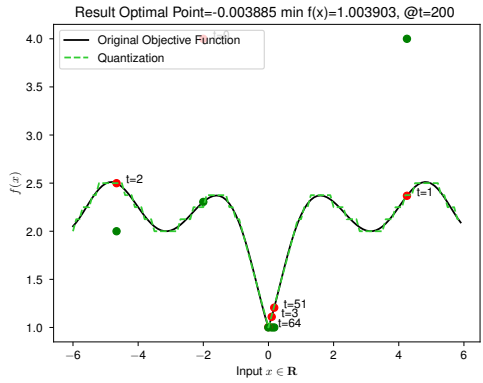}}}\label{fig01-004}
    \hfill
    \subfigure[Search process with respect to Drop wave function]{\resizebox{0.48\textwidth}{!}{\input{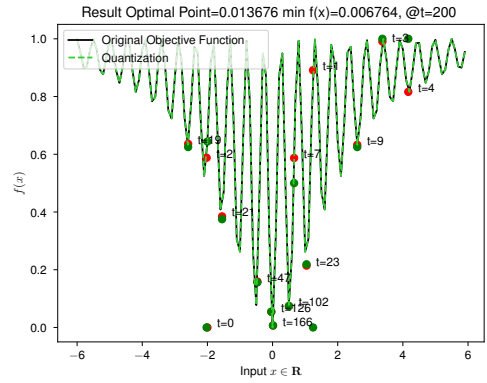}}}\label{fig01-005}
    \hfill
    \subfigure[Search process with respect to Salomon function]{\resizebox{0.48\textwidth}{!}{\input{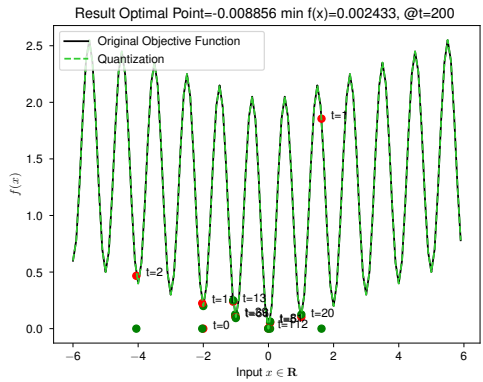}}}\label{fig01-006}
    \hfill
    \subfigure[Search process with respect to Shaffel N2 function]{\resizebox{0.48\textwidth}{!}{\input{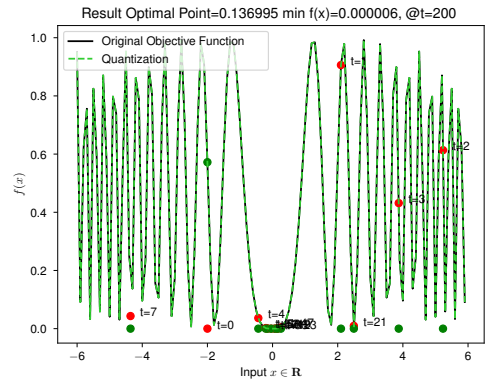}}}\label{fig01-007}
\caption{Search process to benchmark functions on 1-dimension}
\label{fig_tsp_02}
\end{figure}

The benchmark functions used for optimization performance tests are widely known as the complication of finding the global minimum. 
Figure \ref{fig_tsp_01} represents the 3D plot of each benchmark.
All functions involve several local minima and the global minimum at $\bm{x}= \bm{0}$. 
Figure \ref{fig_tsp_02} shows how the quantization-based optimizer searches for the global minimum of the benchmark functions, with the 1D plotted on the Y-zero axis.
Those limited searches on a one-dimensional plane cause the convergence point to be slightly different from the global minimum due to the limited gradient.   
However, as shown in the figures, even if the algorithm is trapped at a local minimum point, the algorithm can escape the local minimum, proceed with searching, and finally converge to the global minimum.

\end{document}